\documentclass[10pt]{article} 
\usepackage[accepted]{tmlr}

\usepackage{amsmath,amsfonts,bm}

\def\eqref#1{(\ref{#1})}

\def\1{\bm{1}}

\DeclareMathAlphabet{\mathsfit}{\encodingdefault}{\sfdefault}{m}{sl}
\SetMathAlphabet{\mathsfit}{bold}{\encodingdefault}{\sfdefault}{bx}{n}

\definecolor{tmlrblue}{rgb}{0.21,0.49,0.74}
\usepackage[pagebackref,breaklinks,colorlinks,linkcolor=tmlrblue,citecolor=tmlrblue]{hyperref}
\usepackage{url}

\usepackage{multirow}
\usepackage{graphicx}
\usepackage{makecell}
\usepackage{xcolor}
\usepackage{colortbl}
\usepackage{hhline}
\usepackage{arydshln}

\usepackage{amssymb}
\newtheorem{theorem}{Theorem}[section]
\newtheorem{proposition}[theorem]{Proposition}

\newtheorem{corollary}[theorem]{Corollary}

\definecolor{DNAgreen}{HTML}{E8F7EA}
\definecolor{rowgray}{HTML}{F5F5F5}

\newtheorem{remark}[theorem]{Remark}
\newenvironment{proof}{\paragraph{Proof.}}{\hfill $\square$}

\title{DNA-Prior: Unsupervised Denoise Anything via\\ Dual-Domain Prior}

\author{\name Yanqi Cheng$^1$, \, Chun-Wun Cheng$^1$, \, Jim Denholm$^2$, \, Thiago Lima$^3$, \\Javier A. Montoya-Zegarra$^{4,3}$, \, Richard Goodwin$^2$, \, Carola-Bibiane Schönlieb$^1$,\\ Angelica I Aviles-Rivero$^5$\thanks{Corresponding Author.}\\ 
\addr $^1$University of Cambridge, $^2$Predictive Artificial Intelligence and Data, Clinical Pharmacology\\ and Safety Sciences, AstraZeneca R\&D, $^3$Lucerne University Teaching and Research Hospital,\\ $^4$Zurich University of Applied Sciences, $^5$YMSC, Tsinghua University }

\begin{document}

\maketitle

\begin{abstract}
Medical image denoising is commonly addressed using supervised models trained on large paired datasets, limiting their applicability across heterogeneous imaging modalities and data-scarce clinical settings. 
To address this limitation, we introduce \textbf{DNA-Prior}, a training-free unsupervised denoising framework that reconstructs clean images directly from corrupted observations by coupling implicit neural and explicit spectral--spatial priors. 
Specifically, an untrained neural parameterisation provides an implicit architectural prior, while a frequency-domain fidelity term and spatial regularisation jointly constrain the reconstruction, requiring neither external training data nor modality-specific training.
Beyond the formulation, we provide a theoretical characterisation of DNA-Prior as a pullback optimisation of the spectral--spatial energy through the neural parameterisation. 
Under a neural tangent kernel approximation and local linearisation of the regulariser, we derive frequency-decoupled reconstruction dynamics that characterise how neural spectral bias and explicit regularisation jointly govern noise suppression and structural preservation, and establish finite optimal stopping behaviour.
Experiments 
demonstrate consistent improvements over classical, implicit-prior, and hybrid unsupervised denoising baselines while preserving fine anatomical structures. 
More importantly, DNA-Prior extends beyond denoising, providing task-agnostic structural guidance to pretrained foundation models without retraining. 
When integrated with Medical-SAM3, it improves downstream segmentation under both box and text prompting, achieving up to an 18.13-percentage-point improvement in Dice score under severe noise and highlighting the potential of structural priors to enhance foundation-model robustness.
\end{abstract}

\section{Introduction}
\label{sec:intro}
High-quality medical images are essential for downstream tasks such as segmentation, registration, and reconstruction, yet clinical acquisitions are often insufficiently constrained, due to the complex characteristics of noise and anatomical structures.  
These degradations can obscure fine anatomical structures and bias quantitative analysis, making effective denoising a critical preprocessing step. Classical methods can suppress noise but often oversmooth important details. Although recent deep learning–based approaches have achieved state-of-the-art performance in CT~\citep{eulig2024benchmarking} and MRI~\citep{zhao2024denoising}, denoising across different modalities remains challenging.
Although supervised denoising—particularly deep learning approaches—have produced strong quantitative gains in CT and MRI, these methods typically depend on large, clean and task-specific training sets whose collection is costly, privacy-constrained, and often impractical. Even when available, models tend to overfit to scanner- or protocol-specific patterns, degrading performance under domain shifts. These issues, combined with heavy computational demands and limited uncertainty estimation, highlight the need for data-efficient denoising approaches that generalise across modalities and noise conditions without requiring paired data~\citep{batson2019noise2self, wang2022blind2unblind}.

Classical explicit priors such as Total Variation~\citep{rudin1992nonlinear}, sparsity~\citep{donoho1995noising}, and smoothness~\citep{tikhonov1977solutions} assumptions have long been used to encode well-understood properties of clean medical and natural images. More recently, implicit priors derived from neural networks — most notably Deep Image Prior (DIP)~\citep{ulyanov2018deep} and self-supervised denoising frameworks like Noise2Self~\citep{batson2019noise2self} — have shown that the architecture of deep models itself can capture powerful structural regularities without paired clean data. 
Recent studies show that integrating explicit and implicit priors improves denoising robustness and generalisation under real-world noise and distribution shifts by uniting handcrafted knowledge with data-driven representations~\citep{zhang2021plug, venkatakrishnan2013plug}.

Beyond reconstruction quality, an important yet underexplored question is whether improved denoising translates into measurable gains in downstream tasks. In clinical practice, it is widely assumed that removing noise enhances segmentation and other high-level analyses; however, this assumption is rarely evaluated in a controlled and model-agnostic manner. In this work, we investigate this connection by examining whether the proposed prior, while designed for denoising, preserves structural information that can be directly exploited by downstream models. This perspective allows us to assess the broader role of denoising as a structural regularisation mechanism rather than a purely preprocessing step.

\textbf{Our contributions are:} (i) We introduce \textbf{DNA-Prior}, a training-free unsupervised denoising framework that couples an implicit neural architectural prior with an explicit spectral--spatial prior, enabling reconstruction directly from a single corrupted observation without external training data.
(ii) We provide a theoretical characterisation of the interaction between these priors. Under an NTK approximation and local linearisation, we derive frequency-decoupled reconstruction dynamics that expose the joint effects of neural spectral bias and explicit regularisation, and establish finite optimal stopping behaviour.
(iii) We demonstrate the generality of DNA-Prior across seven medical imaging modalities and multiple noise levels, consistently outperforming classical, implicit-prior, and hybrid unsupervised baselines while preserving fine anatomical structures.
(iv) We show that DNA-Prior extends beyond image reconstruction by providing task-agnostic structural guidance for downstream segmentation. When integrated with Medical-SAM3 without retraining the foundation model, DNA-Prior improves performance under both box and text prompting, with gains of up to {18.13 Dice} under severe noise, highlighting the potential of structural priors to enhance foundation-model robustness.

\section{Related Work}
The unsupervised denoising method aims to denoise without relying on a clean image, so it does not require a large amount of training data. Non-Local Means (NLM) \citep{buades2005non}, and BM3D \citep{dabov2007image} are traditional unsupervised methods that incorporate image prior knowledge. However, these methods often underperform in complex clinical settings or when handling in vivo noise distributions. Deep unsupervised methods address this limitation by leveraging neural network architectures or learned priors without clean supervision. Deep Image Prior (DIP) \citep{ulyanov2018deep} shows that the structure of a convolutional network itself can serve as an image prior, enabling denoising from a single corrupted image. 
Other methods, such as Recorrupted-to-Recorrupted \citep{pang2021recorrupted}, train networks by generating additional corrupted observations from already noisy images, while score-prior-based variational inference \citep{cheng2023score} leverages learned score priors for real-world single-image denoising. 
Unsupervised strategies have also been explored for low-dose CT denoising using dual-domain information \citep{yu2023unsupervised}. However, such methods often require projection data or reconstruction-related information, which may limit their practical applicability. 

Image denoising has been extensively investigated from the perspective of learning both implicit and explicit image priors. Deep-learning-based methods typically rely on implicit priors, which are learned from large-scale data through network architectures and optimisation strategies. Representative examples include end-to-end residual noise prediction in DnCNN \citep{zhang2017beyond}, the use of noise-level maps in FFDNet \citep{zhang2018ffdnet}, and residual learning with feature-attention mechanisms in RIDNet \citep{anwar2019real}. 
In contrast, classical image priors are generally derived from assumptions about natural image statistics and structures. Examples include total variation regularisation \citep{rudin1992nonlinear}, sparse modeling \citep{mairal2009non}, weighted nuclear norm minimisation based on low-rank matrix approximation \citep{gu2014weighted}, and non-local self-similarity priors employed in methods such as NLM~\citep{buades2005non} and BM3D~\citep{dabov2007image}. Although these explicit priors are often less flexible in handling complex real-world noise, they are effective in preserving image structures and fine details.
Motivated by the complementary strengths of implicit and explicit priors, recent studies have explored hybrid approaches that integrate both types of prior knowledge. For instance, external-prior-guided internal-prior learning combines external statistical information with image-specific priors \citep{xu2018external}, while denoising-prior-driven deep networks incorporate prior knowledge into deep restoration frameworks \citep{dong2018denoising}. 

Denoising has been widely studied to improve downstream segmentation across imaging modalities, particularly in biomedical imaging, where noise is common and expert annotations are costly. Although different modalities suffer from distinct degradations, such as photon noise in microscopy, speckle noise in ultrasound, low-dose noise in CT, and acquisition artifacts in MRI, these disturbances create similar segmentation challenges by obscuring boundaries, reducing foreground–background contrast, and introducing spurious structures. To address these issues, restoration and self-supervised denoising methods, including CARE \citep{weigert2018content}, Noise2Noise \citep{lehtinen2018noise2noise}, Noise2Void \citep{krull2019noise2void}, and Noise2Self \citep{batson2019noise2self}, learn cleaner representations from raw data, which are often easier to obtain than annotated masks. Prior work has shown that denoising can improve segmentation under limited-label settings by preserving task-relevant structures \citep{prakash2020leveraging,liu2017image,wang2019segmentation} and DenoiSeg further demonstrates that jointly optimising denoising and segmentation allows a model to learn from all raw images while using only a few annotated masks \citep{buchholz2020denoiseg}.

\textbf{Denoising for downstream medical image analysis.}
Image degradation can substantially affect downstream medical image analysis by obscuring anatomical boundaries and corrupting task-relevant structures. Prior work has therefore investigated denoising and restoration as preprocessing or jointly learned components for segmentation, including CARE \citep{weigert2018content}, Noise2Noise \citep{lehtinen2018noise2noise}, Noise2Void \citep{krull2019noise2void}, Noise2Self \citep{batson2019noise2self}, and DenoiSeg \citep{buchholz2020denoiseg}. More recently, image enhancement has been studied in conjunction with large pretrained and foundation models~\citep{sun2025foundation}. For example, BME-X demonstrates that learned MRI enhancement can benefit downstream segmentation, registration, and diagnostic tasks, while plug-and-play diffusion-based denoising has been incorporated into segmentation networks for feature refinement.
In contrast, we investigate whether a structural image prior, optimised solely from a corrupted observation, can benefit an already pretrained medical foundation model without retraining or modifying its parameters. Specifically, DNA-Prior operates at the image level and independently of the downstream model, allowing its spectral--spatial prior to provide task-agnostic structural guidance to Medical-SAM3. This setting allows us to examine whether structural information captured during denoising transfers beyond reconstruction and can improve the robustness of downstream foundation-model segmentation under image degradation.

\section{Methodology}
We analyse DNA-Prior as a pullback optimisation of a spatial–spectral energy through a neural denoiser. Using NTK approximation, Fourier diagonalisation, and local regulariser linearisation, we derive frequency-decoupled reconstruction dynamics and establish conditions for convergence and finite optimal stopping.
\begin{figure*}[t!]
  \centering
  \centerline{\includegraphics[width=\linewidth]{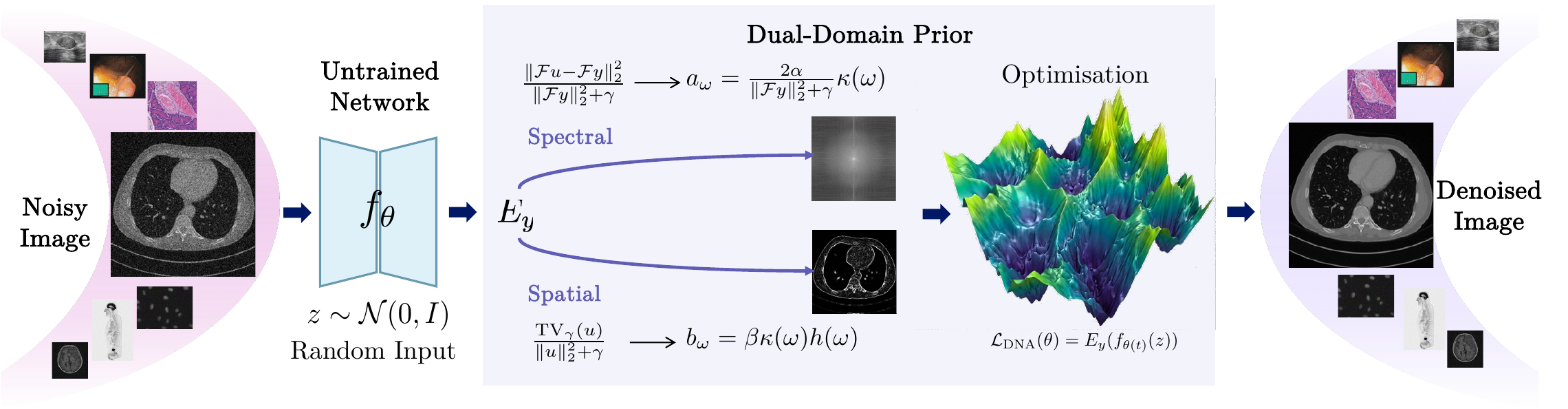}}
\caption{Overview of the denoising flow of DNA-Prior. A random input $z \sim \mathcal{N}(0, I)$ is passed through an untrained network $f_\theta$, then optimised using a dual-domain prior that enforces both spectral and spatial constraints. The noisy image guides the optimisation via the loss $\mathcal{L}_{DNA}(\theta) = E_y(f_{\theta(t)}(z))$ with respect to the noisy observation $y$, producing the final denoised image.}
\label{fig:teaser}
\vspace{-0.2cm}
\end{figure*}

\subsection{Pullback Formulation of DNA-Prior}  We consider the image denoising problem as a special case of the linear
observation model $y = Ax + \varepsilon$, where $x \in \mathbb{R}^n$ is the
unknown clean image, $y \in \mathbb{R}^n$ is the noisy observation,
$A \in \mathbb{R}^{n \times n}$ is the forward operator, and
$\varepsilon \in \mathbb{R}^n$ denotes additive noise. Since this work
focuses on image denoising, we set $A = I$, yielding $y = x + \varepsilon$.
Here, $n = H \times W \times C$, where $H$, $W$, and $C$ denote the image
height, width, and number of channels, respectively.
Following the implicit neural prior framework, we parameterise the
reconstructed image using an untrained neural network $f_\theta$ with a
fixed random input $z \sim \mathcal{N}(0, I)$. The reconstruction at
optimisation time $t$ is defined as $u_t = f_{\theta(t)}(z)$.
To jointly incorporate implicit architectural regularisation and explicit
image-domain constraints, we define the energy functional
\begin{equation}
\label{spatial-spectral}
    E_y(u)
    =
    \alpha
    \frac{\|\mathcal{F}u - \mathcal{F}y\|_2^2}
         {\|\mathcal{F}y\|_2^2 + \gamma}
    +
    \beta
    \frac{\operatorname{TV}_\gamma(u)}
         {\|u\|_2^2 + \gamma},
\end{equation}
where $\mathcal{F}$ denotes the unitary discrete Fourier transform,
$\operatorname{TV}_\gamma$ is a twice continuously differentiable
smoothed total variation regulariser, and $\gamma > 0$ keeps the
denominators strictly positive. The precise discretisation and boundary
conditions of $\operatorname{TV}_\gamma$ are assumed to match the implementation. The parameters
$\alpha, \beta > 0$ control the relative contributions of data fidelity and
spatial regularisation.
Since $\mathcal{F}$ is unitary, Parseval's identity ensures that
$\|\mathcal{F}u - \mathcal{F}y\|_2^2 = \|u - y\|_2^2$.
This equivalence establishes that the Fourier-domain formulation preserves
the original denoising objective without introducing additional assumptions
or altering the data-fidelity measure. Although the fidelity term weights
all frequencies uniformly, the resulting optimisation dynamics may exhibit
frequency-dependent behaviour through the neural parameterisation and
regularisation.

DNA-Prior optimises the pullback of $E_y$ through the neural
parameterisation, yielding the parameter-space objective
\begin{equation}
\label{eq:dna_objective}
    \mathcal{L}_{\mathrm{DNA}}(\theta)
    =
    E_y(f_\theta(z)).
\end{equation}
Along the optimisation trajectory $\theta(t)$, the reconstruction is
$u_t = f_{\theta(t)}(z)$, and hence
$\mathcal{L}_{\mathrm{DNA}}(\theta(t)) = E_y(u_t)$.
To characterise the optimisation dynamics, we consider the continuous-time
gradient flow
\begin{equation}
\label{eq:parameter_flow}
    \frac{d\theta(t)}{dt}
    =
    -\nabla_\theta \mathcal{L}_{\mathrm{DNA}}(\theta(t)).
\end{equation}

Under this formulation, the untrained network provides an implicit
architectural prior, while the explicit energy functional constrains the
reconstruction through data fidelity and spatial regularisation. Their
coupling through the pullback objective induces reconstruction dynamics
that can be analysed in the image space and, under suitable approximations,
in the Fourier domain.

\subsection{Linearised Dynamics and Frequency Decoupling}

Denote the Jacobian as $J_\theta := \frac{\partial u_t}{\partial \theta}$.
For overparameterised networks, the lazy-training approximation assumes
$J_\theta \approx J_0$, and the neural tangent kernel (NTK)
$K_0 = J_0 J_0^\top$ is approximately constant.

Projecting the parameter-space flow onto the image space gives
\begin{equation}
\label{image-space}
\begin{aligned}
\frac{du_t}{dt}
&=
\frac{\partial u_t}{\partial \theta}\frac{d\theta}{dt}
=
J_\theta\bigl(-\nabla_\theta \mathcal{L}_{\mathrm{DNA}}\bigr)
=
J_\theta\left(-J_\theta^\top \nabla_{u_t} E_y(u_t)\right) \\
&\approx
-K_0\left(
\frac{2\alpha}{\|\mathcal{F}y\|_2^2 + \gamma}(u_t - y)
+
\beta \nabla_{u_t} \mathcal{R}(u_t)
\right),
\end{aligned}
\end{equation}
where
$\mathcal{R}(u_t) =
\frac{\operatorname{TV}_\gamma(u_t)}{\|u_t\|_2^2 + \gamma}$.

The image-space dynamics in ~\eqref{image-space} remain difficult
to analyse directly, since the NTK acts as a nonlocal operator and the
regularisation term $\nabla_{u_t}\mathcal{R}(u_t)$ is nonlinear. To obtain
a tractable characterisation, we introduce approximations that allow the
dynamics to be analysed in the Fourier domain.

We first assume periodic boundary conditions and consider the
infinite-width approximation of the network. We further assume that the NTK
is approximately translation invariant. For multichannel images, we
additionally assume that cross-channel interactions are negligible,
allowing a scalar frequency-wise approximation of the NTK. Under these
assumptions, the Fourier basis $\{\psi_\omega\}_{\omega}$ diagonalises
$K_0$, so that
$
K_0 \psi_\omega = \kappa(\omega)\psi_\omega,
$
where $\kappa(\omega) \geq 0$ denotes the frequency-wise NTK spectrum.

We define the Fourier coefficients
$\widehat{u}_t(\omega) = \langle u_t, \psi_\omega \rangle$ and
$\widehat{y}(\omega) = \langle y, \psi_\omega \rangle$.
Projecting the image dynamics onto the Fourier basis yields
\begin{equation}
\begin{aligned}
\frac{d\widehat{u}_t(\omega)}{dt}
&=
-\left\langle
K_0\left(
\frac{2\alpha}{\|\mathcal{F}y\|_2^2 + \gamma}(u_t - y)
+
\beta \nabla_{u_t} \mathcal{R}(u_t)
\right),
\psi_\omega
\right\rangle \\
&=
-a_\omega
\bigl(\widehat{u}_t(\omega) - \widehat{y}(\omega)\bigr)
-
\beta \kappa(\omega)\,
\widehat{\nabla \mathcal{R}(u_t)}(\omega),
\end{aligned}
\end{equation}
where
$a_\omega =
\frac{2\alpha}{\|\mathcal{F}y\|_2^2 + \gamma}\,\kappa(\omega)$.
The quantity $\kappa(\omega)$ determines the effective learning rate
associated with frequency $\omega$. Due to the spectral bias often observed
in convolutional neural networks, $\kappa(\omega)$ may decrease as
$|\omega|$ increases, implying that low-frequency components can evolve
more rapidly during optimisation.
Although the NTK approximation decouples the kernel operator in Fourier
space, the dynamics remain nonlinear because of the regularisation term
$\nabla_{u_t}\mathcal{R}(u_t)$. To obtain analytically tractable local
dynamics, we linearise the regulariser around a stationary reconstruction
$u^*$ of the approximate image-space dynamics, i.e.\ a point satisfying
\begin{equation}
\label{eq:stationary}
K_0 \nabla E_y(u^*) = 0.
\end{equation}
Assuming that the smoothed total variation $\operatorname{TV}_\gamma$ is
twice continuously differentiable for $\gamma > 0$, the regulariser
$\mathcal{R}$ is also twice continuously differentiable. A first-order
Taylor expansion gives
\[
\nabla \mathcal{R}(u_t)
=
\nabla \mathcal{R}(u^*)
+
\nabla^2 \mathcal{R}(u^*)(u_t - u^*)
+
o(\|u_t - u^*\|).
\]

The constant term $\nabla \mathcal{R}(u^*)$ contributes to the equilibrium
condition~\eqref{eq:stationary} but does not affect the leading-order
perturbation dynamics. We therefore analyse the dynamics around $u^*$ by
introducing the perturbation
$
v_t = u_t - u^*,
\quad
\widehat{v}_t(\omega)
=
\widehat{u}_t(\omega) - \widehat{u}^*(\omega).
$
Let $H_{\mathcal{R}} = \nabla^2 \mathcal{R}(u^*)$. The leading-order
behaviour of the regularisation gradient is then characterised by
\begin{equation}
\label{eq:regulariser_linearisation}
\nabla \mathcal{R}(u_t) - \nabla \mathcal{R}(u^*)
\approx
H_{\mathcal{R}}\, v_t.
\end{equation}

For the normalised total variation regulariser, the Hessian includes
contributions from both the smoothed TV numerator and the image-dependent
normalisation denominator. Consequently, its spectral behaviour is not
determined solely by the standard TV operator. To further simplify the
analysis, we adopt a local frozen-coefficient approximation in smooth image
regions, where the spatial coefficients of the linearised operator vary
slowly and may be treated as approximately constant. For the
frequency-decoupled model, we additionally assume that the resulting
linearised operator is approximately translation invariant and
approximately diagonalised by the Fourier basis. This is an additional
modelling assumption, not a general property of normalised TV; its
image-dependent denominator can contribute non-Laplacian and nonlocal
Hessian terms.
Under this approximation, the linearised regulariser acts as a Fourier
multiplier with real-valued symbol $h(\omega)$, such that
$
H_{\mathcal{R}} \psi_\omega \approx h(\omega)\,\psi_\omega.
$
For a constant-coefficient discrete Laplacian surrogate, the
frequency dependence is characterised by
\[
\rho(\omega)
=
4\sin^2\!\left(\frac{\pi\omega_1}{H}\right)
+
4\sin^2\!\left(\frac{\pi\omega_2}{W}\right)
\sim 4\pi^2 |\omega|^2
\]
in the small-normalised-frequency limit, where
$|\omega|^2 = \frac{\omega_1^2}{H^2} + \frac{\omega_2^2}{W^2}$ denotes the
squared normalised frequency magnitude.

We do not in general identify $h(\omega)$ with a positive multiple of
$\rho(\omega)$; such an identification would require an additional
assumption on the local Hessian. In particular, $h(\omega)$ need not be
nonnegative.
Substituting the local linearisation into the projected dynamics and using
the stationary condition~\eqref{eq:stationary} yields the approximate
decoupled frequency-wise evolution equation
\begin{equation}
\frac{d\widehat{v}_t(\omega)}{dt}
=
-\bigl(a_\omega + b_\omega\bigr)\,\widehat{v}_t(\omega), \text{ where, } b_\omega = \beta \kappa(\omega) h(\omega),
\:
a_\omega + b_\omega
=
\kappa(\omega)
\left(
\frac{2\alpha}{\|\mathcal{F}y\|_2^2 + \gamma}
+
\beta h(\omega)
\right).
\label{eq:frequency_dynamics}
\end{equation}
Expression~\eqref{eq:frequency_dynamics} shows that, under the stated
approximations, the local optimisation dynamics approximately decouple
across spatial frequencies. When $a_\omega + b_\omega > 0$, the
corresponding perturbation decays exponentially towards the stationary
reconstruction. The frequency-dependent NTK spectrum governs the learning
rates, while the linearised regularisation operator modifies the stability
and damping of individual frequency components.

\subsection{Convergence Analysis and Stopping Criterion}

Solving the linearised frequency dynamics~\eqref{eq:frequency_dynamics}
yields the closed-form evolution
\begin{equation}
\widehat{v}_t(\omega)
=
e^{-(a_\omega + b_\omega)t}\,\widehat{v}_0(\omega),
\label{eq:solution_frequency}
\end{equation}
so the corresponding reconstruction satisfies
$
\widehat{u}_t(\omega)
=
\widehat{u}^*(\omega)
+
e^{-(a_\omega + b_\omega)t}
\bigl(\widehat{u}_0(\omega) - \widehat{u}^*(\omega)\bigr).
$
This expression describes the local linearised dynamics while the
reconstruction remains sufficiently close to $u^*$.
Decomposing
$\widehat{y}(\omega) = \widehat{x}(\omega) + \widehat{\varepsilon}(\omega)$,
and assuming independent zero-mean Gaussian noise with covariance
$\sigma^2 I$, the unitary Fourier transform gives
$\mathbb{E}[|\widehat{\varepsilon}(\omega)|^2] = \sigma^2$.
We examine the frequency-wise mean squared error relative to the clean
image. Let
\[
\lambda_\omega = a_\omega + b_\omega,
\qquad
c_\omega = \widehat{u}^*(\omega) - \widehat{x}(\omega),
\qquad
d_\omega = \widehat{u}_0(\omega) - \widehat{u}^*(\omega).
\]
Under the additional approximation that $\lambda_\omega > 0$ is fixed
across noise realisations (although the original objective depends on $y$), the frequency-wise mean squared error is
\begin{equation}
\begin{aligned}
\mathrm{MSE}_\omega(t)
&=
\mathbb{E}\left[\left|c_\omega + e^{-\lambda_\omega t} d_\omega\right|^2\right] \\
&=
\underbrace{\mathbb{E}[|c_\omega|^2]}_{\text{asymptotic reconstruction error}}
+
\underbrace{2 e^{-\lambda_\omega t}\,
\operatorname{Re}\mathbb{E}[c_\omega \overline{d_\omega}]}_{\text{transient cross term}}
+
\underbrace{e^{-2\lambda_\omega t}\,
\mathbb{E}[|d_\omega|^2]}_{\text{transient optimisation effects}}.
\end{aligned}
\label{eq:mse_frequency}
\end{equation}
The expectation is taken over the noise and any random initialisation.
Unlike the unnormalised quadratic regularisation case, the stationary
reconstruction $u^*$ may depend nonlinearly on the noisy observation.
Consequently, the asymptotic reconstruction error cannot generally be
expressed as a fixed linear shrinkage of $\widehat{y}(\omega)$.
To characterise the stopping behaviour, let
\[
A_\omega = \mathbb{E}[|d_\omega|^2],
\qquad
B_\omega = \operatorname{Re}\mathbb{E}[c_\omega \overline{d_\omega}].
\]
Then
\[
\mathrm{MSE}_\omega(t)
=
\mathbb{E}[|c_\omega|^2]
+
2 B_\omega e^{-\lambda_\omega t}
+
A_\omega e^{-2\lambda_\omega t}.
\]

\begin{theorem}[Local convergence and finite optimal stopping]
\label{thm:finite_optimal}
Assume that the local frequency-decoupled approximation is used as a
fixed-rate surrogate over the stopping horizon, with decay rate $\lambda_\omega = a_\omega + b_\omega > 0$. Then the
perturbation in frequency mode $\omega$ converges exponentially to zero
within this surrogate model.
If $A_\omega > 0$ and $-A_\omega < B_\omega < 0$, the frequency-wise mean
squared error admits a finite, strictly positive optimal stopping time
\begin{equation}
t_\omega^*
=
\frac{1}{\lambda_\omega}
\log\left(\frac{A_\omega}{-B_\omega}\right).
\label{eq:optimal_time}
\end{equation}
If $A_\omega > 0$ and $B_\omega \leq -A_\omega$, the minimum is attained at
$t_\omega^* = 0$. If $A_\omega > 0$ and $B_\omega \geq 0$, the mean squared
error decreases towards its infimum as $t \to \infty$ (not necessarily
attained at finite time). When $A_\omega = 0$,
the mean squared error is constant in time.
\end{theorem}

\begin{proof}
From~\eqref{eq:solution_frequency},
$\widehat{v}_t(\omega) = e^{-\lambda_\omega t}\widehat{v}_0(\omega)$.
Since $\lambda_\omega > 0$, the perturbation converges exponentially to
zero.
To analyse the stopping time, set $q = e^{-\lambda_\omega t}$, where
$q \in (0, 1]$. Expression~\eqref{eq:mse_frequency} becomes
\[
\mathrm{MSE}_\omega(q)
=
\mathbb{E}[|c_\omega|^2] + 2 B_\omega q + A_\omega q^2.
\]
For $A_\omega > 0$, differentiation gives
$\frac{d\,\mathrm{MSE}_\omega}{dq} = 2 B_\omega + 2 A_\omega q$,
so the unconstrained minimiser is $q_\omega^* = -B_\omega / A_\omega$.

If $-A_\omega < B_\omega < 0$, then $q_\omega^* \in (0, 1)$, and hence
\[
t_\omega^*
=
-\frac{1}{\lambda_\omega}\log q_\omega^*
=
\frac{1}{\lambda_\omega}
\log\left(\frac{A_\omega}{-B_\omega}\right),
\]
which is finite and positive.

If $A_\omega > 0$ and $B_\omega \leq -A_\omega$, the minimum occurs at
$q = 1$, corresponding to $t = 0$. If $A_\omega > 0$ and $B_\omega \geq 0$,
the infimum occurs as $q \to 0$, corresponding to $t \to \infty$. Finally,
if $A_\omega = 0$, then $d_\omega = 0$ almost surely and the mean squared
error is independent of time.
\end{proof}

\begin{remark}[Consistency with DIP early stopping]
\label{rem:dip_limit}
When $\beta = 0$ and the NTK is positive definite on the modes
under consideration, the stationary reconstruction in those modes is
$\widehat{u}^*(\omega)=\widehat y(\omega)$. Then $c_\omega = \widehat{\varepsilon}(\omega)$ and
$d_\omega = \widehat{u}_0(\omega) - \widehat{y}(\omega)$, giving, for deterministic or noise-independent initialisation and fixed
$a_\omega>0$,
$B_\omega=-\sigma^2<0$ and
$A_\omega=\mathbb E[|\widehat{u}_0(\omega)-\widehat{x}(\omega)|^2]+\sigma^2$.
If the initial error has positive expected squared magnitude, then
$A_\omega>\sigma^2$ and Theorem~\ref{thm:finite_optimal} yields
the strictly positive finite stopping time
\[
t_\omega^*
=
\frac{1}{a_\omega}
\log\left(
1 + \frac{\mathbb E[|\widehat{u}_0(\omega) - \widehat{x}(\omega)|^2]}{\sigma^2}
\right),
\]
which exhibits the early-stopping bias--variance trade-off of the
fixed-rate DIP surrogate. If the initial error is zero, the optimum is
$t_\omega^*=0$. Modes with small positive $a_\omega$ evolve slowly;
if $a_\omega=0$, that mode is not updated by the linearised dynamics.
This does not establish the same stopping behaviour for the full nonlinear
network optimisation.
\end{remark}

\begin{corollary}[High-frequency convergence behaviour]
\label{cor:high_freq}
In frequency modes for which the spectral bias of the network gives
$\kappa(\omega) \approx 0$, the local decay rate satisfies
\[
\lambda_\omega
=
\kappa(\omega)
\left(
\frac{2\alpha}{\|\mathcal{F}y\|_2^2 + \gamma}
+
\beta h(\omega)
\right).
\]
Consequently, high-frequency perturbations may converge slowly when
$\lambda_\omega > 0$ is small. The regularisation term modifies this decay
rate through $h(\omega)$, although its effect depends on the combined
frequency dependence of $\kappa(\omega)$ and $h(\omega)$.
\end{corollary}

\begin{proof}
Immediate from the definition of $a_\omega$ and $b_\omega$ and the
closed-form solution~\eqref{eq:solution_frequency}.
\end{proof}

\begin{remark}[Absence of explicit regularisation]
\label{rem:no_reg}
When $\beta = 0$ (no explicit TV regularisation), we have $b_\omega = 0$
and $\lambda_\omega = a_\omega$. Modes satisfying $a_\omega \approx 0$
then evolve arbitrarily slowly in the approximate dynamics. If
$a_\omega = 0$, the corresponding perturbation remains constant in the
linearised model.
\end{remark}

\begin{proposition}[Steady-state reconstruction]
\label{prop:steady_state}
For each frequency mode satisfying $a_\omega + b_\omega > 0$, and
within the regime where the local linearisation remains valid, the local
linearised reconstruction converges to
$
\widehat{u}_\infty(\omega) = \widehat{u}^*(\omega).
$
The convergence rate is governed by $a_\omega + b_\omega$, reflecting the
interaction between the implicit neural prior and the explicit
regularisation term.
\end{proposition}

\begin{proof}
From the closed-form evolution,
$\widehat{u}_t(\omega) = \widehat{u}^*(\omega) +
e^{-(a_\omega + b_\omega)t}
\bigl(\widehat{u}_0(\omega) - \widehat{u}^*(\omega)\bigr)$.
If $a_\omega + b_\omega > 0$, the exponential term vanishes as
$t \to \infty$, giving the stated limit.
\end{proof}

\begin{remark}[Convergence behaviour]
\label{rem:convergence}
The exponentially decaying perturbation shows that stable frequency modes
converge towards the stationary reconstruction in the local model. However,
convergence of the optimisation dynamics does not necessarily imply
monotonic improvement in reconstruction accuracy. Depending on the sign of
the transient cross term in~\eqref{eq:mse_frequency}, the frequency-wise
mean squared error may attain its minimum at a finite stopping time.
\end{remark}

DNA-Prior combines an implicit neural prior, encoded through the NTK
spectrum $a_\omega$, with an explicit variational prior contributing the
term $b_\omega$. Under the idealised assumptions leading to the linearised
frequency-decoupled dynamics, Theorem~\ref{thm:finite_optimal} establishes
local exponential convergence for stable frequency modes and identifies
conditions under which finite optimal stopping occurs.
Corollary~\ref{cor:high_freq} characterises the corresponding
high-frequency convergence behaviour.

\section{Experimental Results}

We evaluate DNA-Prior across seven medical imaging modalities and multiple noise levels, comparing against classical, implicit-prior, and hybrid unsupervised methods using PSNR and qualitative assessments. We further assess its transferability to downstream segmentation with Medical-SAM3 under clean and noisy conditions, demonstrating its effectiveness, robustness, and generalisability.

\subsection{Experimental Setup}
All experiments were conducted on a workstation equipped with a single NVIDIA RTX~4090 GPU (24\,GB VRAM). 
Each denoising task was performed independently for maximum $3{,}000$ iterations per image without any pretraining or external supervision. Specifically, our DNA-Prior uses parameters $\alpha = 1$ and $\beta = 0.0001$.
\begin{table*}[t!]
\centering
\renewcommand{\arraystretch}{1.25}
\resizebox{\linewidth}{!}{
\begin{tabular}{c|c|ccccccccc}
\Xhline{0.25ex}\vspace{0.05cm}
                                 &  & Wavelet & TV    & CBM3D & DIP   & DIP-TV & WIRE  & DS-N2N & DSP   & {\hspace{-0.2cm}{DNA-Prior}}                          \\ 
                                 &\multirow{-2}{*}{$\sigma$}      & (2000)
                                 & (2004)
                                 & (2012)
                                 & (2018)
                                 & (2019)
                                 & (2023)
                                 & (2025)
                                 & (2026)
                                 & {{(2026)}}                          \\ \hline
                                 & 15    & 38.10   & 40.34 & 40.75 & 41.96 & 41.98  & 38.33 & 41.14  & 41.82 & \cellcolor[HTML]{D7FFD7}{ \textbf{42.11}} \\
                                 & \cellcolor[HTML]{EFEFEF}25    & \cellcolor[HTML]{EFEFEF}32.90   & \cellcolor[HTML]{EFEFEF}32.57 & \cellcolor[HTML]{EFEFEF}33.09 & \cellcolor[HTML]{EFEFEF}34.28 & \cellcolor[HTML]{EFEFEF}34.26  & \cellcolor[HTML]{EFEFEF}31.59 & \cellcolor[HTML]{EFEFEF}34.08  & \cellcolor[HTML]{EFEFEF}34.28 & \cellcolor[HTML]{D7FFD7}{ \textbf{34.32}} \\
\multirow{-3}{*}{\begin{tabular}[c]{@{}c@{}}Microscopy\\ ~\citep{caicedo2019nucleus}\end{tabular}}     & 50    & 24.97   & 24.30 & 22.54 & 25.19 & 25.10  & 23.03 & 25.26  & 25.13 & \cellcolor[HTML]{D7FFD7}{ \textbf{25.31}} \\ \hline
                                 & 15    & 30.33   & 31.40 & 32.80 & 33.31 & 33.05  & 33.55 & 32.56  & 33.10 & \cellcolor[HTML]{D7FFD7}{ \textbf{33.69}} \\
                                 & \cellcolor[HTML]{EFEFEF}25    & \cellcolor[HTML]{EFEFEF}26.84   & \cellcolor[HTML]{EFEFEF}27.17 & \cellcolor[HTML]{EFEFEF}29.29 & \cellcolor[HTML]{EFEFEF}29.75 & \cellcolor[HTML]{EFEFEF}29.73  & \cellcolor[HTML]{EFEFEF}29.67 & \cellcolor[HTML]{EFEFEF}28.82  & \cellcolor[HTML]{EFEFEF}29.62 & \cellcolor[HTML]{D7FFD7}{ \textbf{29.76}} \\
\multirow{-3}{*}{\begin{tabular}[c]{@{}c@{}}MRI\\~\citep{msoud_nickparvar_2021}\end{tabular}}            & 50    & 21.93   & 20.07 & 20.85 & 23.56 & 23.70  & 23.99 & 23.05  & 23.69 & \cellcolor[HTML]{D7FFD7}{ \textbf{24.02}} \\ \hline
                                 & 15    & 32.28   & 32.97 & 32.97 & 33.11 & 33.15  & 32.90 & 32.61  & 33.13 & \cellcolor[HTML]{D7FFD7}{ \textbf{33.24}} \\
                                 & \cellcolor[HTML]{EFEFEF}25    & \cellcolor[HTML]{EFEFEF}28.48   & \cellcolor[HTML]{EFEFEF}28.73 & \cellcolor[HTML]{EFEFEF}28.73 & \cellcolor[HTML]{EFEFEF}29.12 & \cellcolor[HTML]{EFEFEF}29.13  & \cellcolor[HTML]{EFEFEF}28.97 & \cellcolor[HTML]{EFEFEF}28.97  & \cellcolor[HTML]{EFEFEF}29.14 & \cellcolor[HTML]{D7FFD7}{ \textbf{29.16}} \\
\multirow{-3}{*}{\begin{tabular}[c]{@{}c@{}}PET\\~\citep{kuangyu_shi_2022_6361846}\end{tabular}}            & 50    & 22.83   & 22.52 & 21.28 & 23.20 & 23.18  & 23.15 & 22.81  & 23.19 & \cellcolor[HTML]{D7FFD7}{ \textbf{23.20}} \\ \hline
                                 & 15    & 28.45   & 29.66 & 30.08 & 30.93 & 30.95  & 30.61 & 31.12  & 30.92 & \cellcolor[HTML]{D7FFD7}{ \textbf{31.38}} \\
                                 & \cellcolor[HTML]{EFEFEF}25    & \cellcolor[HTML]{EFEFEF}25.62   & \cellcolor[HTML]{EFEFEF}26.64 & \cellcolor[HTML]{EFEFEF}28.96 & \cellcolor[HTML]{EFEFEF}29.25 & \cellcolor[HTML]{EFEFEF}29.18  & \cellcolor[HTML]{EFEFEF}28.77 & \cellcolor[HTML]{EFEFEF}28.43  & \cellcolor[HTML]{EFEFEF}29.21 & \cellcolor[HTML]{D7FFD7}{ \textbf{29.32}} \\
\multirow{-3}{*}{\begin{tabular}[c]{@{}c@{}}CT\\~\citep{albertina2016cancer}\end{tabular}}             & 50    & 21.16   & 19.89 & 20.49 & 23.97 & 23.84  & 24.12 & 23.70  & 23.96 & \cellcolor[HTML]{D7FFD7}{ \textbf{24.57}} \\ \hline
                                 & 15    & 32.07   & 31.24 & 31.78 & 25.25 & 32.42  & 32.70 & 32.83  & 32.42 & \cellcolor[HTML]{D7FFD7}{ \textbf{32.92}} \\
                                 & \cellcolor[HTML]{EFEFEF}25    & \cellcolor[HTML]{EFEFEF}29.08   & \cellcolor[HTML]{EFEFEF}30.41 & \cellcolor[HTML]{EFEFEF}29.16 & \cellcolor[HTML]{EFEFEF}26.23 & \cellcolor[HTML]{EFEFEF}26.97  & \cellcolor[HTML]{EFEFEF}30.32 & \cellcolor[HTML]{EFEFEF}30.33  & \cellcolor[HTML]{EFEFEF}30.48 & \cellcolor[HTML]{D7FFD7}{ \textbf{30.77}} \\
\multirow{-3}{*}{\begin{tabular}[c]{@{}c@{}}Endoscopy\\~\citep{jha2019kvasir}\end{tabular}}      & 50    & 24.41   & 23.85 & 20.38 & 23.18 & 25.49  & 25.17 & 24.87  & 25.43 & \cellcolor[HTML]{D7FFD7}{ \textbf{25.50}} \\ \hline
                                 & 15    & 28.98   & 28.92 & 27.28 & 30.01 & 29.80  & 28.77 & 30.72  & 30.05 & \cellcolor[HTML]{D7FFD7}{ \textbf{31.22}} \\
                                 & \cellcolor[HTML]{EFEFEF}25    & \cellcolor[HTML]{EFEFEF}26.57   & \cellcolor[HTML]{EFEFEF}28.01 & \cellcolor[HTML]{EFEFEF}26.61 & \cellcolor[HTML]{EFEFEF}29.47 & \cellcolor[HTML]{EFEFEF}29.49  & \cellcolor[HTML]{EFEFEF}28.10 & \cellcolor[HTML]{EFEFEF}28.80  & \cellcolor[HTML]{EFEFEF}29.45 & \cellcolor[HTML]{D7FFD7}{ \textbf{30.11}} \\
\multirow{-3}{*}{Histopathology} & 50    & 23.87   & 23.60 & 25.11 & 26.75 & 26.70  & 25.64 & 25.71  & 26.84 & \cellcolor[HTML]{D7FFD7}{ \textbf{27.13}} \\ \hline
                                 & 15    & 31.52   & 33.28 & 33.42 & 34.38 & 34.43  & 31.89 & 34.81  & 34.78 & \cellcolor[HTML]{D7FFD7}{ \textbf{36.54}} \\
                                 & \cellcolor[HTML]{EFEFEF}25    & \cellcolor[HTML]{EFEFEF}29.20   & \cellcolor[HTML]{EFEFEF}30.98 & \cellcolor[HTML]{EFEFEF}31.55 & \cellcolor[HTML]{EFEFEF}32.97 & \cellcolor[HTML]{EFEFEF}34.05  & \cellcolor[HTML]{EFEFEF}31.33 & \cellcolor[HTML]{EFEFEF}32.45  & \cellcolor[HTML]{EFEFEF}33.29 & \cellcolor[HTML]{D7FFD7}{ \textbf{34.40}} \\
\multirow{-3}{*}{\begin{tabular}[c]{@{}c@{}}Ultrasound\\~\citep{al2020dataset}\end{tabular}}     & 50    & 26.33   & 24.47 & 27.75 & 30.17 & 30.02  & 28.44 & 28.85  & 30.07 & \cellcolor[HTML]{D7FFD7}{ \textbf{30.27}} \\ \Xhline{0.25ex}
\end{tabular}
}
\caption{Comparison of our proposed DNA-Prior on PSNR (dB) with classical and unsupervised baselines, including Wavelet \citep{chang2000adaptive}, TV \citep{chambolle2004algorithm}, CBM3D \citep{lebrun2012analysis}, DIP \citep{ulyanov2018deep}, DIP-TV \citep{liu2019image}, WIRE \citep{saragadam2023wire}, DS-N2N \citep{bai2025dual}, and DSP \citep{cheng2025deep}, all tested under identical noise settings at $\sigma = 15, 25, 50$.}
\label{tab:quant_results}
\end{table*}

\textbf{Datasets and Noise Settings.}
To evaluate the generalisation ability of the proposed method, 
we conducted experiments on seven representative medical imaging modalities:
Microscopy~\citep{caicedo2019nucleus}, 
MRI~\citep{msoud_nickparvar_2021}, 
PET~\citep{kuangyu_shi_2022_6361846}, 
CT~\citep{albertina2016cancer}, 
Endoscopy~\citep{jha2019kvasir}, Histopathology (in-house dataset), and 
Ultrasound~\citep{al2020dataset}. 
For each modality, synthetic Gaussian noise with standard deviations $\sigma = 15, 25, 50$ 
was added to simulate different degradation levels. 
Only noisy images were used during optimisation, while clean images were reserved for quantitative evaluation.

\begin{figure*}[t!]
  \centering
  \centerline{\includegraphics[width=0.85\linewidth]{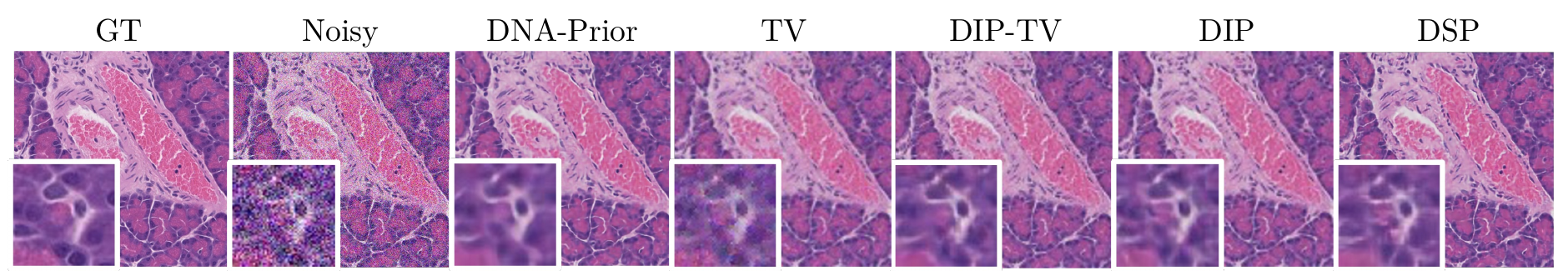}\vspace{-0.2cm}}
\caption{Comparison of denoising results at $\sigma = 50$ on histopathology: DNA-Prior versus existing state-of-the-art methods. Our comparison spans explicit-prior (TV), implicit-prior (DIP, DSP), and hybrid (DIP-TV) approaches. Zoomed-in views are included to enable detailed evaluation.
}
\label{fig:histopathology}
\vspace{-0.2cm}
\end{figure*}

\begin{figure*}[t!]
  \centering
  \centerline{\includegraphics[width=0.85\linewidth]{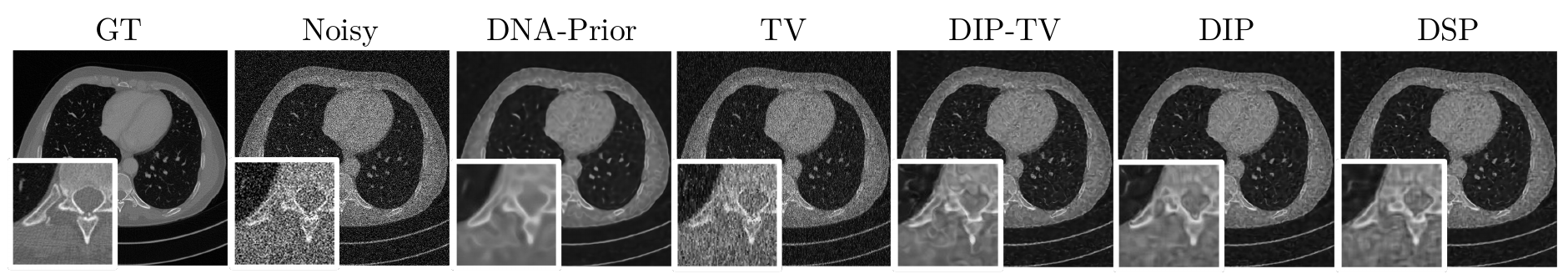}\vspace{-0.2cm}}
\caption{Denoising performance at $\sigma = 50$ for computed tomography (CT), contrasting DNA-Prior against representative explicit-prior (TV), implicit-prior (DIP, DSP), and hybrid (DIP-TV) methods, with Magnified detailed views provided.}
\label{fig:CT}
\vspace{-0.2cm}
\end{figure*}
\begin{figure*}[t!]
  \centering
  \centerline{\includegraphics[width=0.85\linewidth]{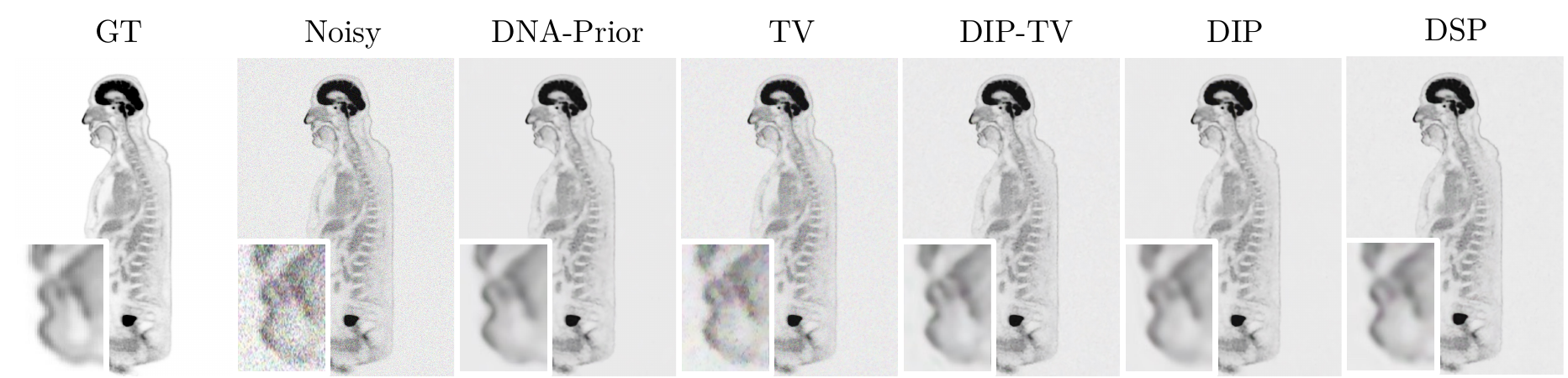}\vspace{-0.2cm}}
\caption{Visual evaluation of denoising results at $\sigma = 50$ for positron emission tomography (PET), comparing DNA-Prior with top-performing established methods. Enlarged views are shown to highlight structural details along edge regions.}
\label{fig:PET}
\vspace{-0.2cm}
\end{figure*}
\begin{figure*}[t!]
  \centering
  \centerline{\includegraphics[width=0.85\linewidth]{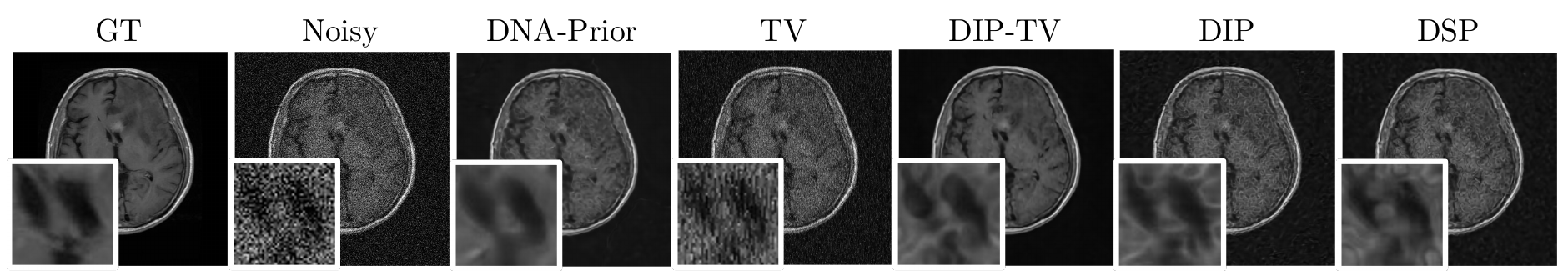}\vspace{-0.2cm}}
\caption{Visual comparison of denoising outcomes at $\sigma = 50$ on magnetic resonance imaging (MRI), positioning DNA-Prior against current state-of-the-art techniques. Zoomed-in regions are presented to allow close inspection of texture fidelity and artifact reduction.}
\label{fig:MRI}
\vspace{-0.2cm}
\end{figure*}

\textbf{Evaluation Metric.}
The denoising performance was assessed using the Peak Signal-to-Noise Ratio (PSNR), which is the mean squared error between the denoised output and the clean ground truth. 
Higher PSNR values indicate better denoising quality and higher fidelity to the original image.

\subsection{Results and Discussion}
\begin{figure*}[!t]
  \centering
  \centerline{\includegraphics[width=0.85\linewidth]{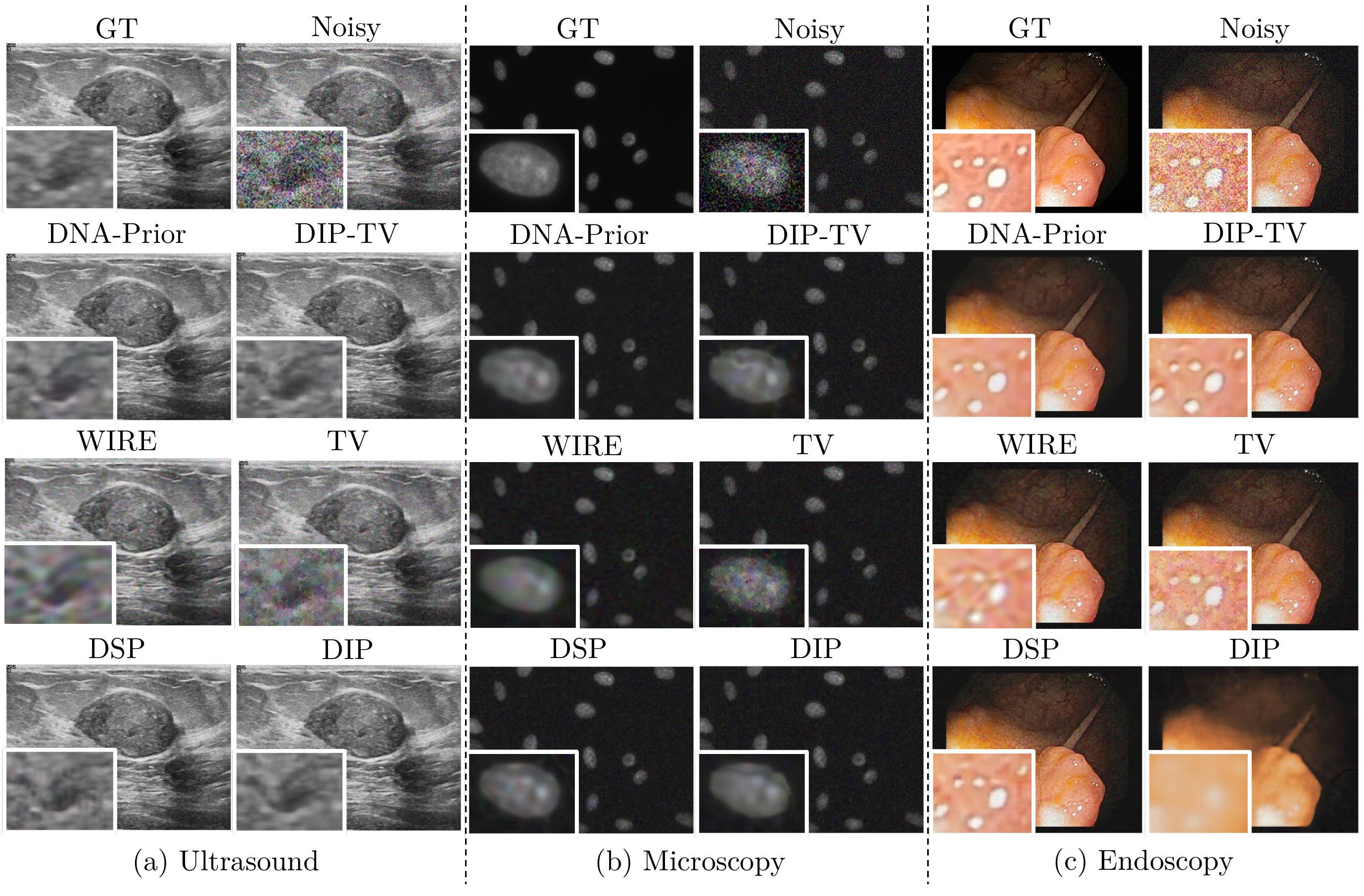}\vspace{-0.3cm}}
\caption{Denoising comparison at $\sigma = 50$ on (a) ultrasound, (b) microscopic and (c) endoscopic imaging, benchmarking DNA-Prior against explicit-prior (TV), implicit-prior (DIP, DSP, WIRE), and hybrid (DIP-TV) baselines. Zoomed-in regions are provided for detailed visual assessment. 
}
\label{fig:ultrasound-microscopy-endoscopy}
\vspace{-0.5cm}
\end{figure*}

Quantitative and qualitative results across all modalities and noise levels are reported in Table~\ref{tab:quant_results}, and Fig.~\ref{fig:histopathology},~\ref{fig:CT},~\ref{fig:MRI},~\ref{fig:PET},~\ref{fig:ultrasound-microscopy-endoscopy} respectively. 
Methods relying purely on explicit priors, such as Wavelet~\citep{chang2000adaptive}, TV~\citep{chambolle2004algorithm} and CBM3D\citep{lebrun2012analysis}, 
achieve reasonable noise suppression but tend to oversmooth fine anatomical structures due to their limited modeling capacity. 
Implicit-prior methods, including DIP~\citep{ulyanov2018deep}, DSP~\citep{cheng2025deep} and WIRE~\citep{saragadam2023wire}, 
better preserve high-frequency textures by exploiting image-specific structural regularisation, 
yet introduce residual artifacts or inconsistent contrast in complex regions. 
Hybrid techniques, such as DIP+TV~\citep{liu2019image}, and DS-N2N~\citep{bai2025dual}, leading to more balanced reconstructions under complex imaging scenario across all 7 modalities. 
Our proposed method, which further integrates an adaptive explicit regulariser with a spectral implicit prior, 
consistently achieves the highest PSNR across all datasets. 
Notably, substantial improvements are observed for CT and Ultrasound modalities, 
where balancing smoothness and detail preservation is particularly challenging.

\textbf{Histopathology.} DNA-Prior achieves the best PSNR performance across all noise levels, consistently outperforming both classical and hybrid approaches. From the visual comparison in Fig.~\ref{fig:histopathology}, residual noise is still shown in TV-based reconstructions, while competing implicit and hybrid methods exhibit grid-like artefacts and twisted structural patterns. In contrast, DNA-Prior produces cleaner reconstructions with substantially fewer artefacts. These results demonstrate the robustness of the proposed method in handling highly textured and structurally complex data, which is characteristic of histopathology images.
\textbf{Computed Tomography (CT).} DNA-Prior achieves the highest PSNR across all noise levels, with performance gains higher under severe noise, indicating strong robustness in challenging conditions. Qualitatively, as shown in Fig.~\ref{fig:CT}, explicit-prior methods such as TV and DIP-TV tend to produce over-smoothed, “plastic-like” reconstructions, while implicit approaches like DIP and DSP preserve edges but introduce speckled artefacts and noise leakage. In contrast, DNA-Prior effectively balances high-frequency detail preservation with noise suppression, yielding sharper edge reconstruction and maintaining structural fidelity in regions such as bone interfaces. This results in more clinically plausible tissue appearance, highlighting the advantage of the proposed dual-domain prior in CT imaging. 
\begin{table*}[]
\centering
\renewcommand{\arraystretch}{1.2}
\resizebox{\linewidth}{!}{
\begin{tabular}{c|c|cccc|cccc}
\Xhline{0.25ex}\vspace{0.05cm}
 & w/wo & \multicolumn{4}{c|}{Box Prompt} & \multicolumn{4}{c}{Text Prompt} \\ \cline{3-10}
 & DNA-Prior & Dice (\%) & IoU (\%) & PSNR & SSIM & Dice (\%) & IoU (\%) & PSNR & SSIM \\ \hline

\multirow{3}{*}{\begin{tabular}[c]{@{}c@{}}CVC-Clinic\\~\citep{bernal2017comparative}\end{tabular}} 
& w/o
& 91.53 & 85.60 & 21.08 & 90.05 
& 87.90 & 81.13 & 20.53 & 85.06 \\

& w/
& \cellcolor[HTML]{D7FFD7}{\textbf{91.57}} 
& \cellcolor[HTML]{D7FFD7}{\textbf{85.67}} 
& \cellcolor[HTML]{D7FFD7}{\textbf{21.10}} 
& \cellcolor[HTML]{D7FFD7}{\textbf{90.11}} 
& \cellcolor[HTML]{D7FFD7}{\textbf{87.99}} 
& \cellcolor[HTML]{D7FFD7}{\textbf{81.19}} 
& \cellcolor[HTML]{D7FFD7}{\textbf{20.54}} 
& \cellcolor[HTML]{D7FFD7}{\textbf{85.12}} \\

& $\Delta$
& +0.04 & +0.07 & +0.02 & +0.06
& +0.09 & +0.06 & +0.01 & +0.06 \\ \hline

\multirow{3}{*}{\begin{tabular}[c]{@{}c@{}}CHASE\_DB1\\~\citep{fraz2012ensemble}\end{tabular}} 
& w/o 
& 32.55 & 20.87 & 5.93 & 17.89 
& 62.51 & 45.59 & 11.89 & 54.71 \\

& w/
& \cellcolor[HTML]{D7FFD7}{\textbf{50.68}} 
& \cellcolor[HTML]{D7FFD7}{\textbf{36.21}} 
& \cellcolor[HTML]{D7FFD7}{\textbf{10.67}} 
& \cellcolor[HTML]{D7FFD7}{\textbf{40.67}} 
& \cellcolor[HTML]{D7FFD7}{\textbf{64.68}} 
& \cellcolor[HTML]{D7FFD7}{\textbf{47.90}} 
& \cellcolor[HTML]{D7FFD7}{\textbf{11.90}} 
& \cellcolor[HTML]{D7FFD7}{\textbf{54.94}} \\ 

& $\Delta$
& +18.13 & +15.34 & +4.74 & +22.78
& +2.17 & +2.31 & +0.01 & +0.23 \\ 

\Xhline{0.25ex}
\end{tabular}
}
\caption{Comparison between Medical-SAM3 with (w/) and without (w/o) DNA-Prior under Box and Text prompts on CVC-Clinic and CHASE\_DB1 datasets, evaluated using Dice (\%), IoU (\%), PSNR, and SSIM. The improvement ($\Delta$) provided by DNA-Prior is shown for each metric. Best results are highlighted in green.}
\label{tab:sam3_results}
\end{table*}
\begin{figure*}[!t]
  \centering
  \centerline{\includegraphics[width=0.85\linewidth]{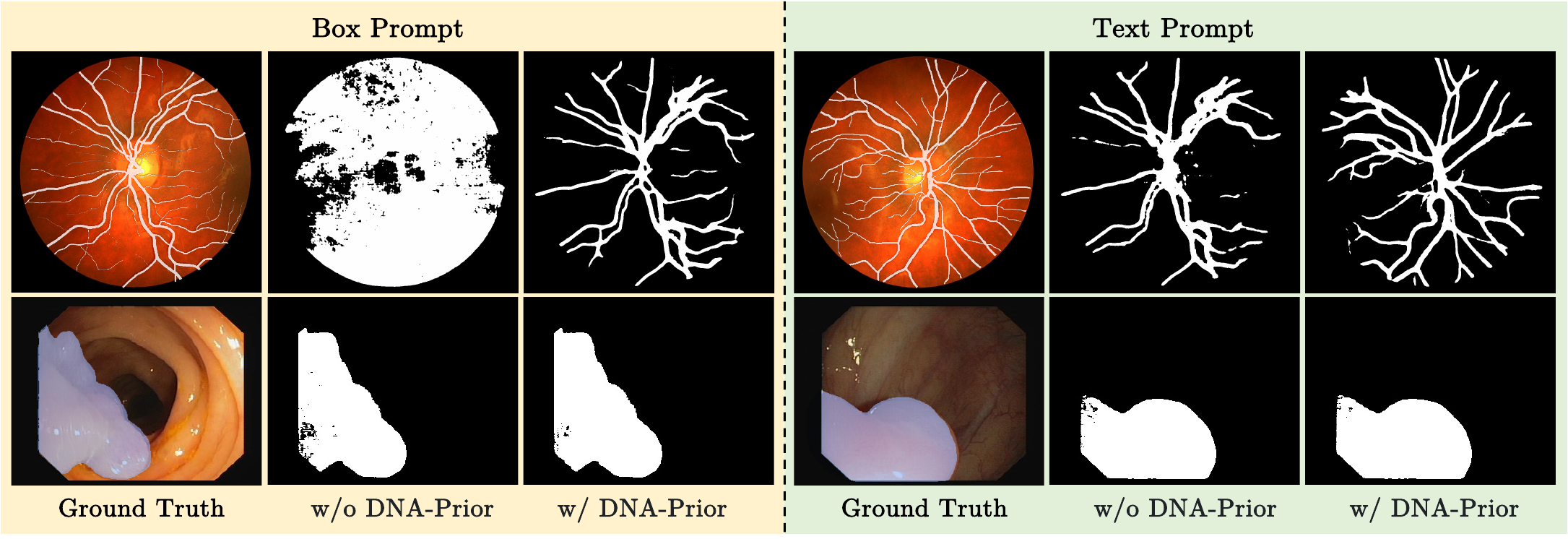}}
\caption{Visual comparison of segmentation results on the CHASE\_DB1 (top) and CVC-Clinic (bottom) datasets on Medical-SAM3~\citep{jiang2026medicalsam3} w/wo the DNA-Prior denoising. Box prompts are automatically generated from ground truth masks as bounding box coordinates. Text prompts use dataset-specific natural language descriptions: ``Retinal Blood Vessel'' for CHASE\_DB1, ``Polyp'' for CVC-ClinicDB.}\vspace{-0.5cm}
\label{fig:seg_results}
\end{figure*}
\textbf{Magnetic Resonance Imaging (MRI).} DNA-Prior outperforms both explicit and implicit baselines in terms of PSNR, with a moderate performance gap at low noise levels that widens as noise increases, reflecting improved robustness to noise variability. Explicit-prior methods degrade significantly under high noise, while implicit-prior approaches exhibit instability due to overfitting to noise patterns. Qualitative comparisons in Fig.~\ref{fig:MRI} show that explicit methods tend to lose anatomical sharpness, particularly along soft tissue boundaries, whereas implicit methods better preserve textures but introduce artefacts such as unnatural intensity variations. In contrast, DNA-Prior achieves a balanced reconstruction by preserving tissue interfaces while effectively suppressing noise without introducing distortions, demonstrating the benefit of integrating spectral and spatial priors.
\textbf{Positron Emission Tomography (PET).} For PET imaging, DNA-Prior achieves the highest PSNR across all tested noise levels, with consistent gains over both classical and deep baselines. Although the improvement margin is smaller compared to other modalities, it remains stable, reflecting reliable performance in data with inherently lower spatial resolution. As shown in Fig.~\ref{fig:PET}, reconstructions from explicit-prior methods exhibit noticeable residual noise, while implicit approaches partially recover detail but tend to over-smooth boundaries. In contrast, DNA-Prior effectively suppresses noise while preserving smooth yet structurally consistent regions, avoiding artificial patterns and maintaining physiologically meaningful signal distributions.
\textbf{Ultrasound.} Ultrasound exhibits one of the largest improvements with DNA-Prior, particularly at lower noise levels, achieving the highest PSNR across all noise settings. As shown in Fig.~\ref{fig:ultrasound-microscopy-endoscopy}, explicit-prior methods retain noticeable speckle artefacts, while WIRE introduces colour bubble artefacts and other implicit methods produce structural distortions. In contrast, DNA-Prior effectively suppresses noise in ultrasound while preserving anatomical structures and texture consistency.
\textbf{Microscopy.} Across all noise levels, DNA-Prior consistently achieves the highest PSNR compared to all baselines. Classical explicit-prior methods (Wavelet, TV, CBM3D) show clear performance degradation as noise increases, indicating limited robustness to complex noise distributions. Implicit-prior approaches (DIP, DSP) remain competitive at low noise but fail to maintain performance under severe corruption, a limitation that is particularly evident in WIRE. Hybrid methods such as DIP-TV provide improved stability but still fall short of DNA-Prior, highlighting the benefit of the proposed dual-domain constraint in high-noise microscopy settings. Qualitative results from Fig.~\ref{fig:ultrasound-microscopy-endoscopy} further show that explicit-prior methods tend to over-smooth images, leading to loss of fine cellular textures, while implicit methods preserve high-frequency details but introduce structural inconsistencies. In contrast, DNA-Prior preserves both global structural integrity and fine-grained details, resulting in clearer cell boundaries and more realistic texture reconstruction.
\textbf{Endoscopy.} The improvement margin of DNA-Prior on endoscopic data is particularly notable compared to purely implicit methods, which struggle to model the complex, texture-rich patterns characteristic of this modality. Hybrid approaches yield competitive performance but remain consistently below DNA-Prior across both low and high noise regimes. As illustrated in Fig.~\ref{fig:ultrasound-microscopy-endoscopy}, explicit methods such as TV leave noticeable residual noise, while WIRE introduces twisted and unnatural patterns, and DIP tends to over-smooth fine surface textures. In contrast, DNA-Prior produces more realistic reconstructions by preserving intricate mucosal textures while effectively removing noise, resulting in visually coherent and clinically meaningful outputs.

In summary, the quantitative and qualitative evaluations across all seven medical imaging modalities consistently demonstrate that DNA-Prior outperforms existing explicit, implicit, and hybrid denoising methods under varying noise levels. While classical explicit-priors suffer from oversmoothing and implicit approaches often introduce artifacts or instability, our proposed framework uniquely integrates an adaptive explicit regulariser with a spectral implicit prior. This dual-domain strategy achieves superior noise suppression without compromising structural fidelity, as evidenced by the highest PSNR scores across all datasets and visually coherent reconstructions that preserve textures and boundaries.

\subsection{Cross-task Transfer: Segmentation Task}
We further show that the proposed prior improves downstream segmentation performance when integrated with models such as Medical-SAM3, supporting its role as a general structural regulariser.
We follow the experimental setting of Medical-SAM3~\citep{jiang2026medicalsam3} and evaluate the effect of DNA-Prior on two datasets with distinct noise characteristics: CVC-Clinic~\citep{bernal2017comparative}, which contains relatively clean images, and CHASE\_DB1~\citep{fraz2012ensemble}, which exhibits significantly higher noise levels. This setup allows us to assess whether the proposed denoising prior benefits downstream segmentation across varying noise conditions.

\textbf{Quantitative Results.}
As shown in Table~\ref{tab:sam3_results}, integrating DNA-Prior consistently improves segmentation performance across both datasets and prompting strategies (box and text). On the relatively clean CVC-Clinic dataset, the improvements are modest but consistent, indicating that DNA-Prior does not degrade performance even when noise is limited. In contrast, on the noisy CHASE\_DB1 dataset, the gains are substantial, with Dice improving from 32.55 to 50.68 (+18.13) under box prompts and similar trends observed across IoU, PSNR, and SSIM. This demonstrates that DNA-Prior significantly enhances robustness of the segmentation model under severe noise conditions, suggesting that the learned prior effectively removes noise while preserving task-relevant structures.

\textbf{Qualitative Results.}
Visual results in Fig.~\ref{fig:seg_results} further support these findings. On CVC-Clinic, segmentation outputs with and without DNA-Prior are visually similar, though the denoised inputs lead to slightly cleaner and less noisy segmentation mask. However, on CHASE\_DB1, the difference is observed clearly: without DNA-Prior, predictions are noisy, and fail to capture fine vascular structures especially only with box prompt, whereas applying DNA-Prior yields more continuous, accurate, and structurally coherent segmentations. In particular, thin vessel branches are better preserved and false positives are reduced, indicating improved structural awareness. These observations confirm that DNA-Prior acts as an effective structural regulariser, improving the quality of feature representations fed into the foundation model.
Overall, the results demonstrate that DNA-Prior not only improves denoising quality but also translates into meaningful downstream gains, especially in challenging noisy scenarios, validating its role as a general and task-agnostic preprocessing module for medical image segmentation.

\section{Conclusion}
We introduced DNA-Prior, a universal unsupervised denoising framework combining an implicit architectural prior with explicit spectral–spatial regularisation.
The resulting learning formulation reconstructs clean images directly from corrupted measurements while preserving frequency structure and anatomical detail.
Experiments across multiple modalities demonstrate stable and generalisable performance, supporting its utility in medical imaging workflows. Future work includes investigating the performance of the method under different types of clinically relevant, modality-specific noise.

\subsubsection*{Acknowledgments}
AIAR gratefully acknowledges the Tsinghua-PolyU Joint Research Initiative Fund and the Tsinghua University Dushi Program.  This work is also supported by the Yau Mathematical Sciences Center, Tsinghua University.
YC is funded by an AstraZeneca studentship and a Google studentship. 
CWC, JAMZ and AIAR acknowledge support from the Swiss National Science Foundation (SNSF) under grant number 20HW-1 220785.  
CBS acknowledges support from the Philip Leverhulme Prize, the Royal Society Wolfson Fellowship, the EPSRC advanced career fellowship EP/V029428/1, EPSRC grants EP/S026045/1 and EP/T003553/1, EP/N014588/1, EP/T017961/1, the Wellcome Innovator Awards 215733/Z/19/Z and 221633/Z/20/Z, the European Union Horizon 2020 research and innovation programme under the Marie Skodowska-Curie grant agreement No. 777826 NoMADS, the Cantab Capital Institute for the Mathematics of Information and the Alan Turing Institute. 

\bibliography{main}
\bibliographystyle{tmlr}


\end{document}